\documentclass[conference,a4paper]{IEEEtran}
\IEEEoverridecommandlockouts
\usepackage{cite}
\usepackage{amsmath,amssymb,amsfonts}
\usepackage{algorithmic}
\usepackage{romannum}
\usepackage{graphicx}
\usepackage{textcomp}
\usepackage{xcolor}
\def\BibTeX{{\rm B\kern-.05em{\sc i\kern-.025em b}\kern-.08em
    T\kern-.1667em\lower.7ex\hbox{E}\kern-.125emX}}
\begin{document}

\title{UAS Navigation in the Real World Using Visual Observation}

\author{\IEEEauthorblockN{Yuci Han, Jianli Wei, Alper Yilmaz,}

\IEEEauthorblockA{\textit{Photogrammetric Computer Vision Lab., The Ohio State University, Columbus, OH, USA} \\
{\{han.1489, wei.909, yilmaz.15\}@osu.edu}\\
}
}

\maketitle

\begin{abstract}

This paper presents a novel end-to-end Unmanned Aerial System (UAS) navigation approach for long-range visual navigation in the real world. Inspired by dual-process visual navigation system of human's instinct: environment understanding and landmark recognition, we formulate the UAS navigation task into two same phases. Our system combines the reinforcement learning (RL) and image matching approaches. First, the agent learns the navigation policy using RL in the specified environment. To achieve this, we design an interactive UASNAV environment for the training process. Once the agent learns the navigation policy, which means 'familiarized themselves with the environment', we let the UAS fly in the real world to recognize the landmarks using image matching method and take action according to the learned policy. During the navigation process, the UAS is embedded with single camera as the only visual sensor. We demonstrate that the UAS can learn navigating to the destination hundreds meters away from the starting point with the shortest path in the real world scenario.

\end{abstract}

\begin{IEEEkeywords}
deep reinforcement learning, image matching, visual navigation
\end{IEEEkeywords}

\section{Introduction}

UAS Navigation in a GPS-denied environment is a very challenging cognitive task almost out of reach in the past few decades. With the development of modern deep learning, recent research has demonstrated the agent's visual navigation ability without GPS in the virtual simulation environment such as AI2THOR \cite{Zhu2017TargetdrivenVN} and Gibson \cite{Xia_2018_CVPR}. Unfortunately, the real world environment is more complicated and challenging due to illumination variation, seasonal difference and construction changes, which are always consistent in the virtual environment. Since the visual navigation strongly relies on robust visual feature representation of the environment, it makes the real world UAS navigation even harder. 

\begin{figure}[htbp]
\centerline{\includegraphics[width=3.5in]{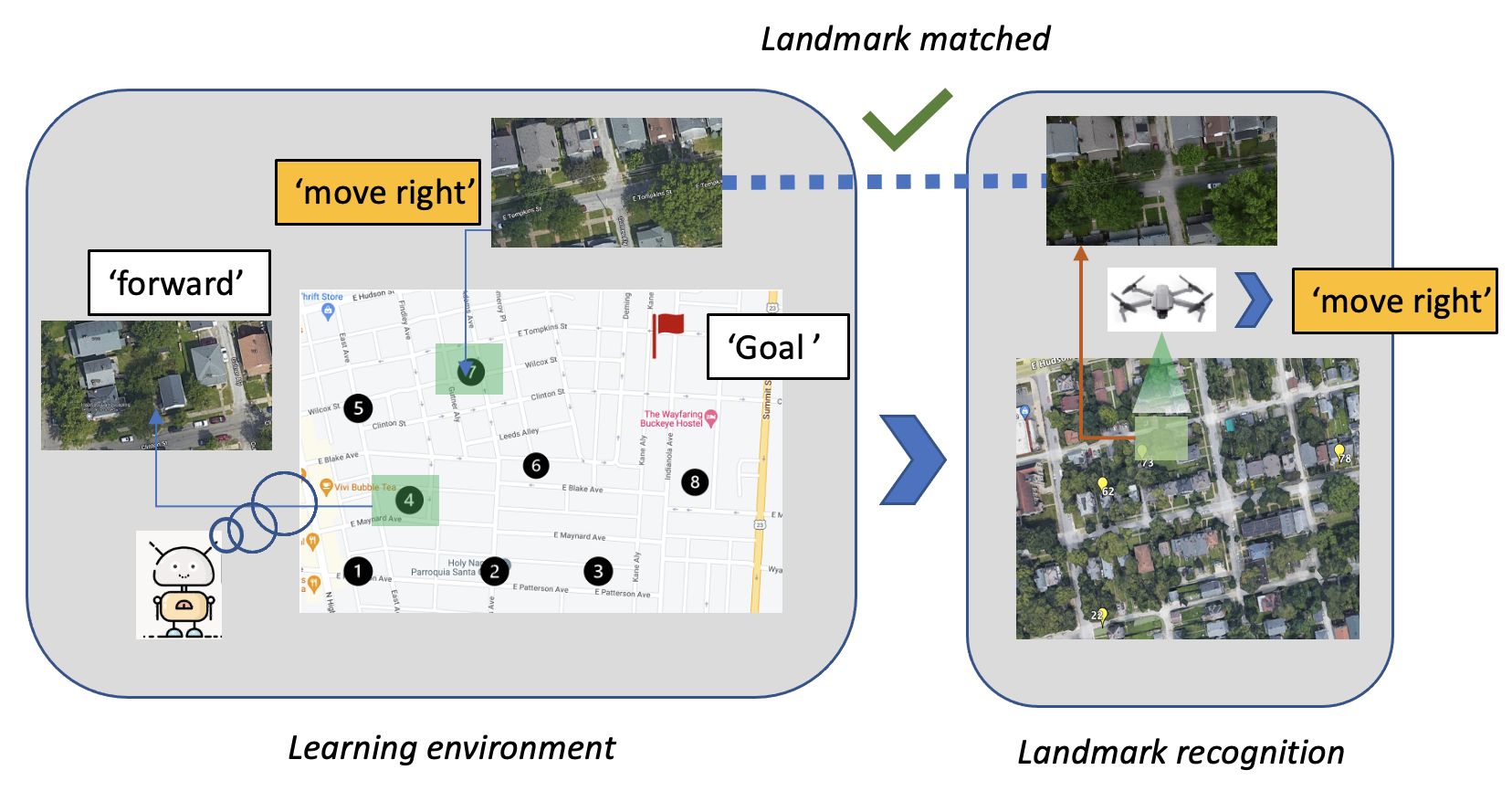}}
\caption{Two phases of our UAS navigation system. The agent learns the navigation policy using RL during the training process. While navigating in the real world, the agent recognize the landmarks and take action based on the learned policy.}
\label{fig1}
\end{figure}

In this work, we propose an architecture that integrates the reinforcement learning and image matching methods that can effectively address the challenging real world UAS navigation task using only visual observations without knowing the map. This is inspired by human's navigation capability in a familiar environment. Imaging that we are walking following the guidance on the GoogleMap to a target place near our house, even though we lose the GPS and are not able to localize our position on the map, we can still find the way by recognizing the landmarks that we know and take action based on what we see to reach to the destination. Our navigation system is developed in two phases (see Fig.1). First, we make the UASNAV environment and train the agent to learn the navigation policy by interacting with the environment using reinforcement learning. During the learning process, the agent familiarizes itself with the predefined landmarks in this environment and generates the navigation policy based on these landmarks to find a shortest way to the target destination. Then, we validate the navigation ability of our agent in the real scenario using UAS. While flying at a certain level, the UAS continuously observes the environment with a down-looking view camera. We adopt SuperGlue \cite{sarlin20superglue} algorithm to match the observed image with our predefined landmarks, even though the features at the same location always change over time due to illumination variance or moving vehicles, we can always match two images sharing similar texture. Once the observation matches with the landmark, the UAS takes actions based on learned policy such as move forward, backward, left or right until reaching the goal point. Our experiment shows the agent's ability to navigate to the target location hundreds meters away.

The contributions of this paper are (a) to present a RL framework for the UAS navigation task, (b) to develop a real scenario interactive environment for the learning process, and (c) to test the proposed approach in the real world using image matching for landmark recognition.

\section{Related work}

\subsection{Visual Navigation}

Visual navigation research in the virtual environment has shown promising results to traverse to the target with the shortest trajectory \cite{Zhu2017TargetdrivenVN}, \cite{Bruce2017OneShotRL}, \cite{kulhanek2021visual}. Some recent work focus on introducing memory to the navigation task, such as graph attention memory (GAM) based system \cite{Li2019GraphAM} and visual graph memory (VGM) structure to embed its navigation history information \cite{Kwon_2021_ICCV}. However, the majority of work in this field is dedicated to use simulation environment and lack validation in the real world .

\subsection{Image Matching}

Image matching as a fundamental concept in computer vision and pattern recognition is the general name of keypoint detection, description and matching. The goal is to recognize the same item in images taken from any angles, with any lighting and scale. Image matching are generally categorized as handcrafted and deep learning. Handcrafted image matching requires expert acknowledge to model local keypoint texture regardless of rotation or scale variation. Over past decades, growing amount of methods have been proposed \cite{ma2021image}. SIFT and SURF \cite{lowe2004distinctive, bay2008speeded} are pieces of the art among them. Nowadays, many applications are still using them. Apart from handcrafted methods, deep descriptors are the main stream in recent years. Those descriptors keep the same vector structure but learnt by a deep learning framework. Paul-Edouard \textit{et al.} proposed SuperGlue \cite{sarlin20superglue}, an end-to-end framework to find matched features between two or more images.

\section{Methodology}

Our proposed method consists of the policy learning and environment recognition phases. This section introduces the RL model, image matching model and the UASNAV environment.

\subsection{Policy Learning using Deep Reinforcement Learning}

We formulate the navigation task as a Markov decision process (MDP) and implement the RL approach. The RL algorithm we adopt is the DCQN model proposed by \cite{2021ISPAn..51..145H}. The goal of the agent is to exploit a policy that rewards reaching the destination starting from any location. The navigation process contains a sequence of states, actions and rewards $(s_0, a_0, r_0, s_1, ... , a_{N-1}, r_{N-1}, s_N)$. The state in this task is the UAS's observation which is a RGB image. We make it a policy that the UAS makes decision on where to go while encountering the landmark. The actions are moving forward, backward, left and right. There are three situations for reward design: 0.1 upon reaching the goal, -0.001 while colliding and -0.0001 for time penalty to encourage finding the shortest trajectory. The agent continuously interacts in the environment to explore the optimal navigation policy by trial and error and maximize the accumulated reward.

\subsection{UASNAV Environment}

The UASNAV environment is designed for training the agent in the policy learning phase. This environment spans a 400 meters by 300 meters area of residential block. We select 100 landmarks distributed in a 10 by 10 pattern in this area. The horizontal spacing is 40 meters and the vertical spacing is 30 meters between each landmark. These landmarks act as ground control points guiding the UAS to the goal. We collect satellite images from Google Maps at these locations to represent the landmark observation. This image is used as the input to the RL algorithm. The output of the DRL module is the optimal action the agent should take at this location (see Fig.2).   The raw image resolution is $1280\times720$, we use the $640\times480$ resized image in the image matching phase.

\begin{figure}[htbp]
\centerline{\includegraphics[width=3.5in]{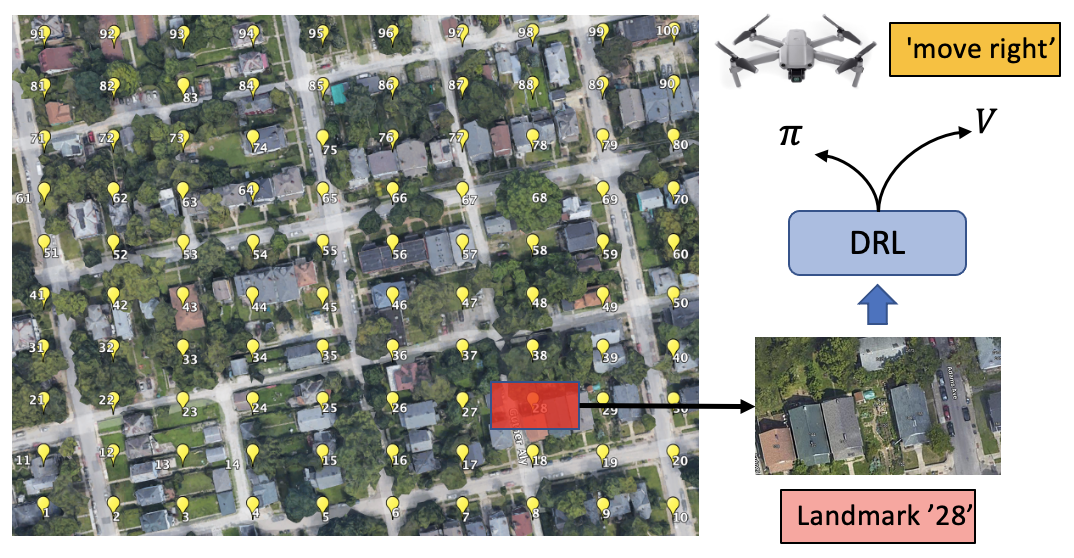}}
\caption{The UASNAV environment spans a 400 meters by 300 meters residential area. We select 100 landmarks and get corresponding satellite image representations. This image is used as the input to the RL algorithm. The output of the DRL module is the optimal action the agent should take at this location} 
\label{fig2}
\end{figure}

\subsection{Environment Recognition using Image Matching}

In the UASNAV environment, we have the satellite image as a descriptor corresponded to each landmark. While navigating in the real environment, the UAS takes real-time observation using the embedded downward-looking camera in a certain frequency. In order to make decision on actions at these ground control point based on the learned policy, we need to enable the UAS to recognize the landmark during the navigation process. Since the UAS observation and satellite image at the same landmark position would contain similar texture. We therefore utilized image matching approach to match the environment landmark descriptor (satellite image) and UAS observation (real-time taken image).  

\begin{figure}[htbp]
\centerline{\includegraphics[width=3.5in]{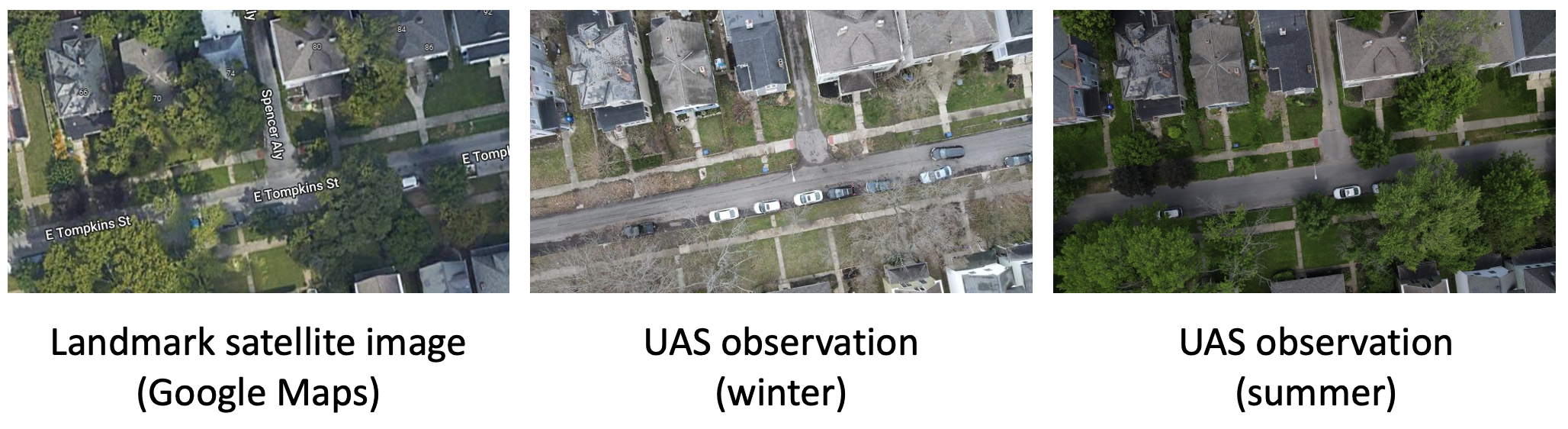}}
\caption{The landmark satellite image and UAS observations in different seasons} 
\label{fig3}
\end{figure}

During the navigation process, the UAS continuously matches the current observation with four landmark (front, back, left, right) descriptors. We choose the one with the most matching points as the matched landmark. To ensure the UAS exactly arrived at the matched landmark position, we use affine matrix to calculate the center distance between the current observation and landmark satellite image. Once it has the shortest center distance, we consider the UAS has arrived at that landmark position and should take action correspondingly.

We adopt SuperGlue \cite{sarlin20superglue}, which is a state of the art algorithm for image matching. It has robust performance with regard to illumination variation or seasonal changes (see Fig.3). Our method takes this advantage, therefore, shows the ability to resist environment variation and is applicable in different seasons and most weather conditions.

\section{Experiments}

We conduct the experiment in two phases. We first train the agent in UASNAV environment to learn the navigation policy. Then, to validate our approach, we operate the UAS to do the goal reaching task in the real world environment to demonstrate its navigation capability without knowing the map.

\subsection{Navigation Policy Learning}

We implement the DCQN \cite{2021ISPAn..51..145H} model for policy learning and observe that the agent successfully learns the navigation task. The training progress in Fig.4 shows the fast convergence of our algorithm. The reward in this figure indicates the cumulative reward computed at each episode as the agent reaches the destination. As shown in Fig.4, the agent learns the navigation policy within 200 episodes. We also run evaluation for 100 episodes in UASNAV environment and demonstrate that the agent is able to find the target in 6.53 steps in average.  

\begin{figure}[!htbp]
\centerline{\includegraphics[width=3.5in]{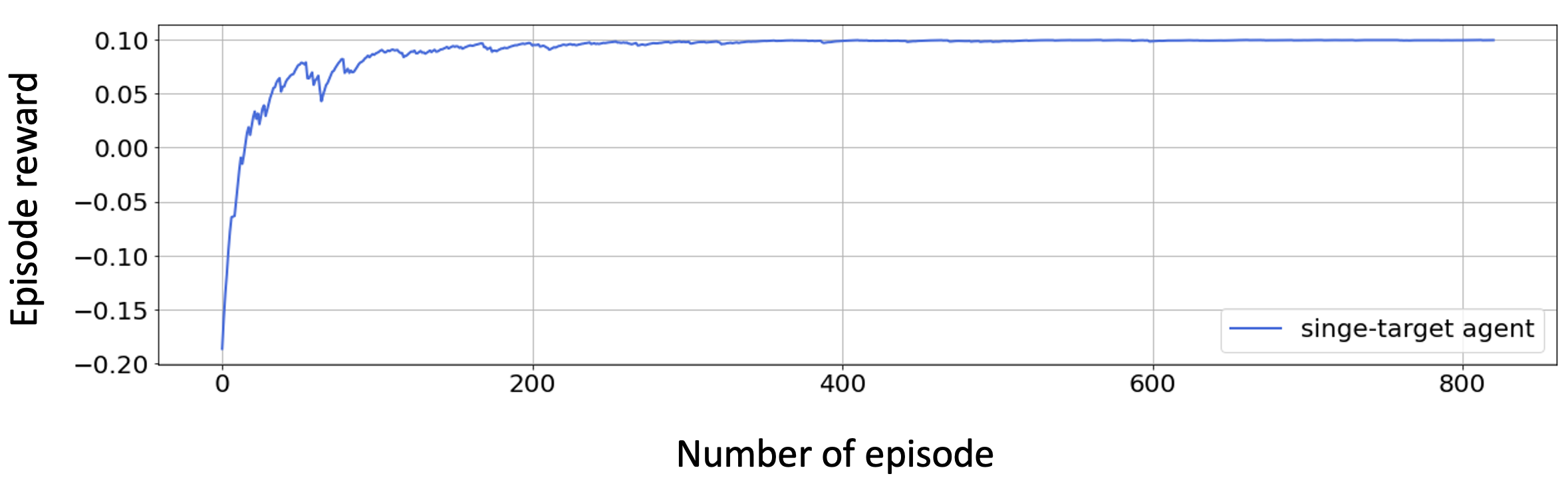}}
\caption{Training curve tracking the agent's episode reward} 
\label{fig4}
\end{figure}

% \begin{figure}[!htbp]
% \centering
% {\includegraphics[width=1in]{fig5.png}%
% \label{fig5}}
% \hfil
% {\includegraphics[width=1in]{fig6.png}%
% \label{fig6}}
% \hfil
% {\includegraphics[width=1in]{fig7.png}%
% \label{fig7}}
% \caption{Illustration of sample trajectories of envalutation result in the UAVNAV environment}
% \end{figure}

\subsection{Real World Navigation}

For the real world test, we use UAS to fly in the residential area which is the same as the UASNAV environment. The UAS is first set to a known starting point and then uses only a down-looking camera to get observation during the whole process. Once it matches with the landmark, the drone will choose an action to either fly forward, backward, left or right based on the output policy of RL algorithm. We adopt the pretrained SuperGlue algorithm for the purpose of image matching in our approach without additional training. The UAS trajectory and image matching results at the landmark location are shown in Fig.5. 

\begin{figure}[!htbp]
\centering
{\includegraphics[width=2.5in]{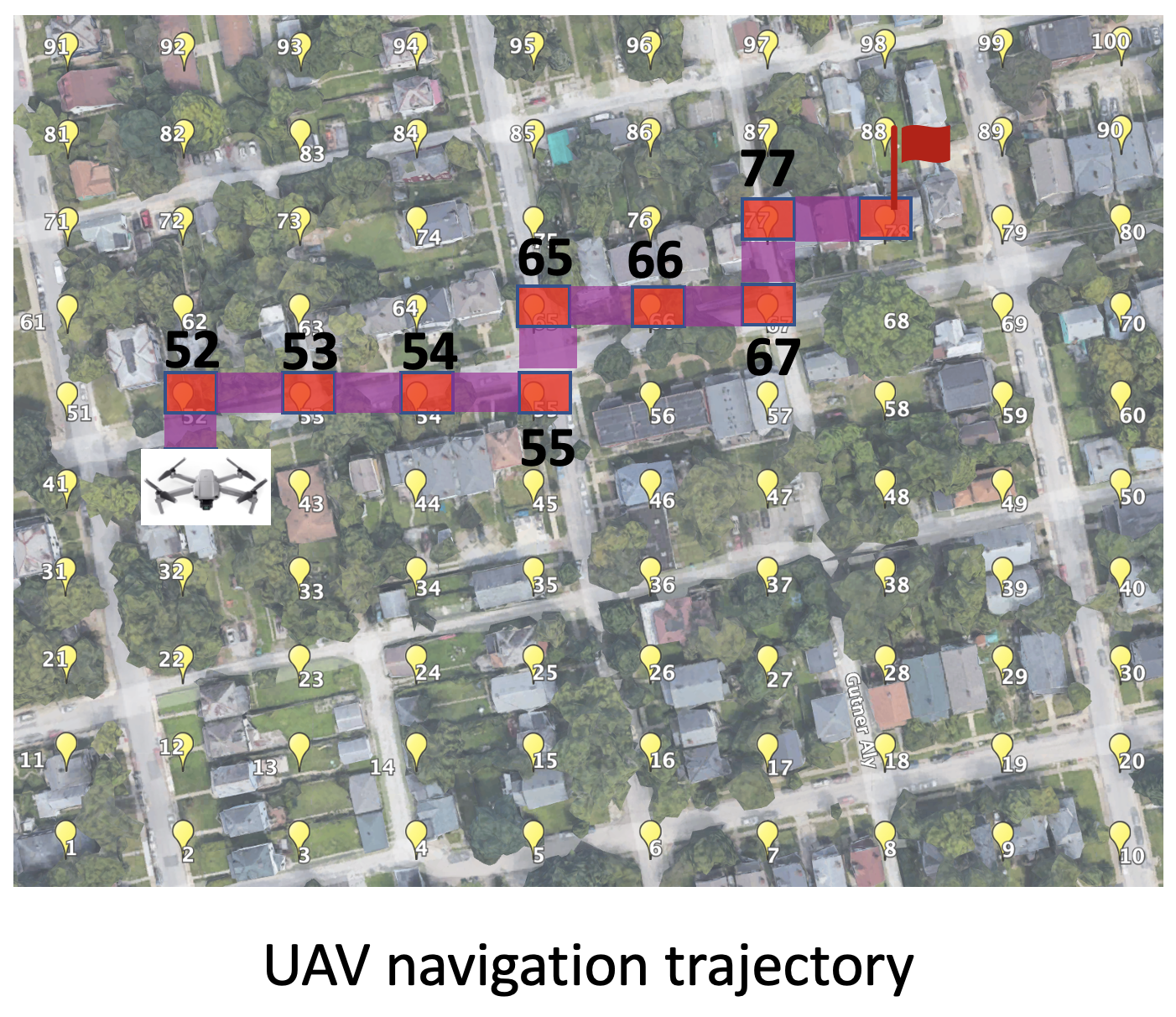}%
\label{fig8}}
{\includegraphics[width=3in]{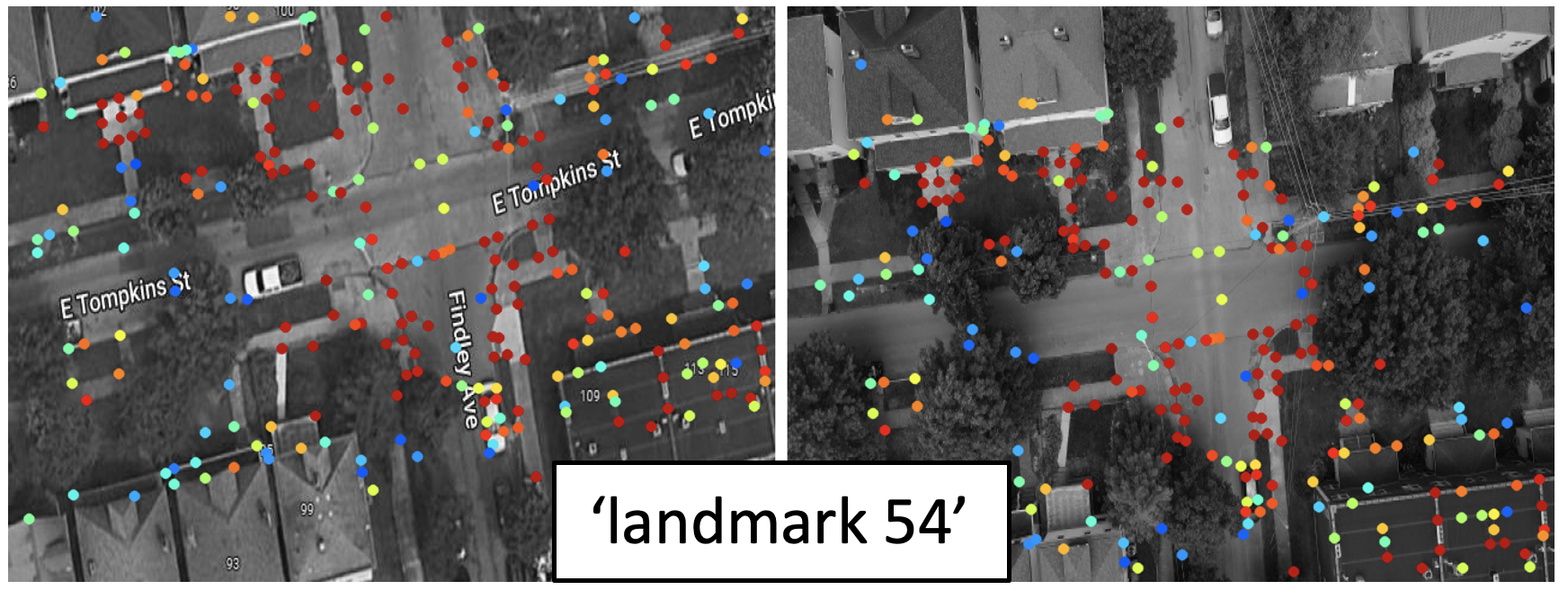}%
\label{fig11}}
{\includegraphics[width=3in]{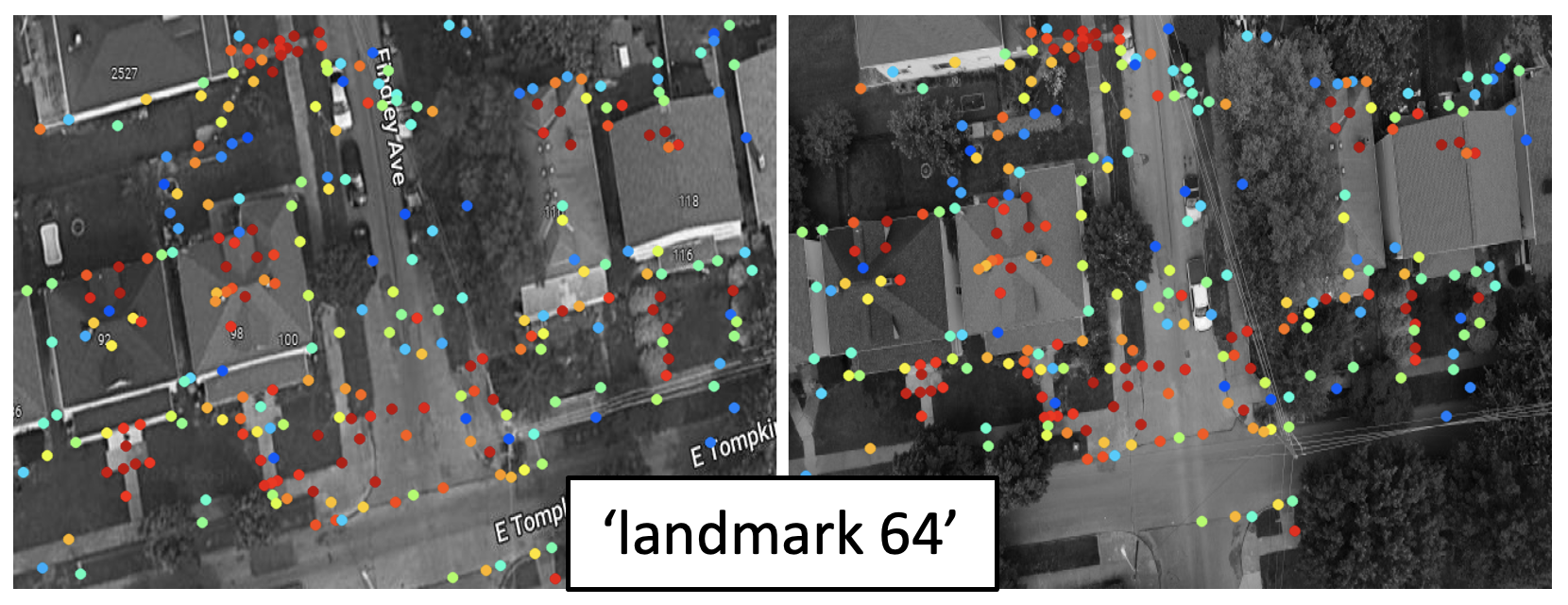}%
\label{fig12}}
{\includegraphics[width=3in]{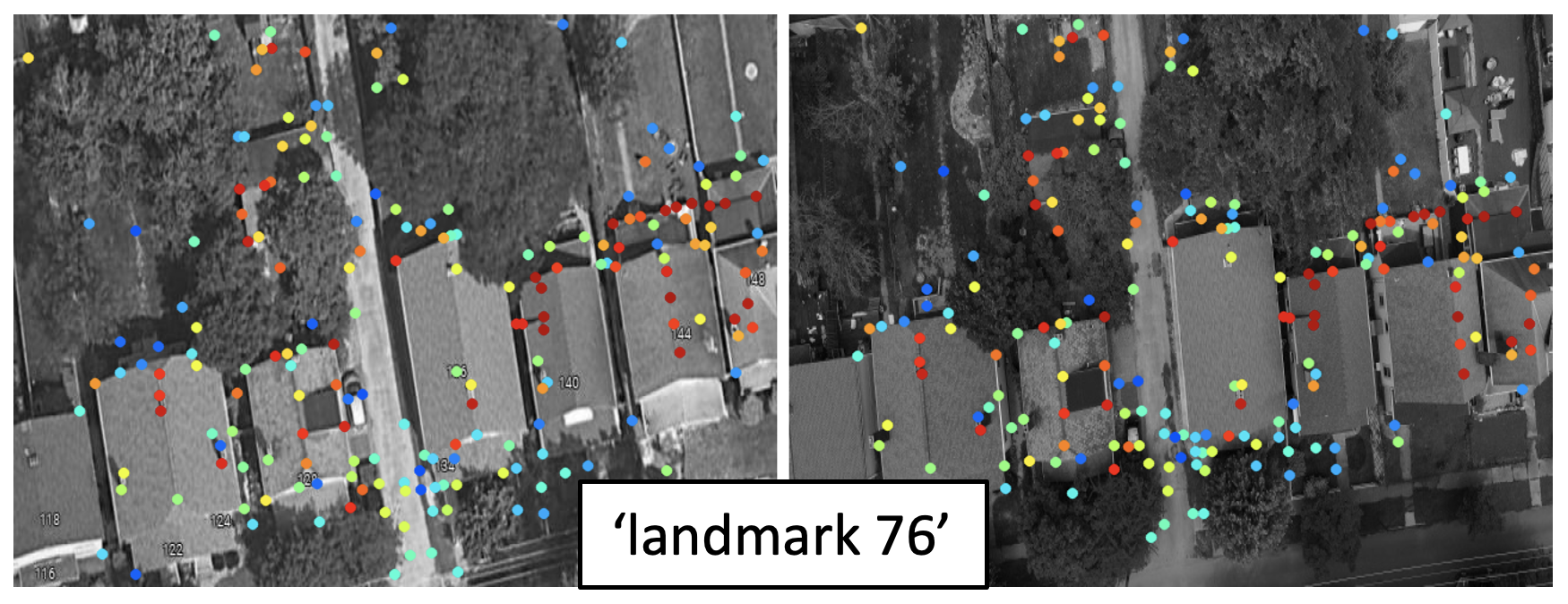}%
\label{fig15}}
\caption{UAS navigation trajectory in the real environment and image matching results at example landmark locations (landmark satellite image vs. UAS observation)}
\end{figure}

% With a NVIDIA TITAN \Romannum{5} GPU, the running speed of our algorithm is approximately \textit{0.42} second per frame, which enables our algorithm to be used for real-time application. 

\section{Conclusion}

We propose a dual phases system for UAS navigation in the real world environment. We formulate the navigation task as MDP and implement RL approach for policy learning. During the real application, we introduce the 'ground control point' concept and adopt image matching method for landmark recognition. Our approach achieves UAS navigation with shortest path without GPS support in an end-to-end manner and has robust performance with regard to the scene changes due to season and weather variation. We demonstrate the UAS's ability to reach the goal hundreds meters away. While conducting the real test, we notice that the UAS has only four action choices at each landmark location, which is inflexible and disables the shortest path to a straight line. The UAS is set to be north oriented during the test, which somehow limits its real application. Besides, our approach is applicable to larger scale environment. To achieve that, we need to set more landmarks, therefore, the limit actions will increase the flying distance which is inefficient. Also, adding more landmarks makes the policy learning process time-consuming. We will tackle these problems by enlarging the action space and optimize the RL algorithm in the future work.

\section{Acknowledgements}

This work is supported by the funding provided from the US Army Office of Research grant AWD-110906.

\bibliographystyle{IEEEtran}
\bibliography{IEEEexample}

\begin{thebibliography}{10}
\providecommand{\url}[1]{#1}
\csname url@samestyle\endcsname
\providecommand{\newblock}{\relax}
\providecommand{\bibinfo}[2]{#2}
\providecommand{\BIBentrySTDinterwordspacing}{\spaceskip=0pt\relax}
\providecommand{\BIBentryALTinterwordstretchfactor}{4}
\providecommand{\BIBentryALTinterwordspacing}{\spaceskip=\fontdimen2\font plus
\BIBentryALTinterwordstretchfactor\fontdimen3\font minus
  \fontdimen4\font\relax}
\providecommand{\BIBforeignlanguage}[2]{{%
\expandafter\ifx\csname l@#1\endcsname\relax
\typeout{** WARNING: IEEEtran.bst: No hyphenation pattern has been}%
\typeout{** loaded for the language `#1'. Using the pattern for}%
\typeout{** the default language instead.}%
\else
\language=\csname l@#1\endcsname
\fi
#2}}
\providecommand{\BIBdecl}{\relax}
\BIBdecl

\bibitem{Zhu2017TargetdrivenVN}
Y.~Zhu, R.~Mottaghi, E.~Kolve, J.~J. Lim, A.~K. Gupta, L.~Fei-Fei, and
  A.~Farhadi, ``Target-driven visual navigation in indoor scenes using deep
  reinforcement learning,'' \emph{2017 IEEE International Conference on
  Robotics and Automation (ICRA)}, pp. 3357--3364, 2017.

\bibitem{Xia_2018_CVPR}
F.~Xia, A.~R. Zamir, Z.~He, A.~Sax, J.~Malik, and S.~Savarese, ``Gibson env:
  Real-world perception for embodied agents,'' in \emph{Proceedings of the IEEE
  Conference on Computer Vision and Pattern Recognition (CVPR)}, June 2018.

\bibitem{sarlin20superglue}
P.-E. Sarlin, D.~DeTone, T.~Malisiewicz, and A.~Rabinovich, ``{SuperGlue}:
  Learning feature matching with graph neural networks,'' in \emph{CVPR}, 2020.

\bibitem{Bruce2017OneShotRL}
J.~Bruce, N.~S{\"u}nderhauf, P.~W. Mirowski, R.~Hadsell, and M.~Milford,
  ``One-shot reinforcement learning for robot navigation with interactive
  replay,'' \emph{ArXiv}, vol. abs/1711.10137, 2017.

\bibitem{kulhanek2021visual}
J.~Kulh{\'a}nek, E.~Derner, and R.~Babu{\v{s}}ka, ``Visual navigation in
  real-world indoor environments using end-to-end deep reinforcement
  learning,'' \emph{IEEE Robotics and Automation Letters}, vol.~6, no.~3, pp.
  4345--4352, 2021.

\bibitem{Li2019GraphAM}
D.~Li, D.~Zhao, Q.~Zhang, Y.~Zhuang, and B.~Wang, ``Graph attention memory for
  visual navigation,'' \emph{ArXiv}, vol. abs/1905.13315, 2019.

\bibitem{Kwon_2021_ICCV}
O.~Kwon, N.~Kim, Y.~Choi, H.~Yoo, J.~Park, and S.~Oh, ``Visual graph memory
  with unsupervised representation for visual navigation,'' in
  \emph{Proceedings of the IEEE/CVF International Conference on Computer Vision
  (ICCV)}, October 2021, pp. 15\,890--15\,899.

\bibitem{ma2021image}
J.~Ma, X.~Jiang, A.~Fan, J.~Jiang, and J.~Yan, ``Image matching from
  handcrafted to deep features: A survey,'' \emph{International Journal of
  Computer Vision}, vol. 129, no.~1, pp. 23--79, 2021.

\bibitem{lowe2004distinctive}
D.~G. Lowe, ``Distinctive image features from scale-invariant keypoints,''
  \emph{International journal of computer vision}, vol.~60, no.~2, pp. 91--110,
  2004.

\bibitem{bay2008speeded}
H.~Bay, A.~Ess, T.~Tuytelaars, and L.~Van~Gool, ``Speeded-up robust features
  (surf),'' \emph{Computer vision and image understanding}, vol. 110, no.~3,
  pp. 346--359, 2008.

\bibitem{2021ISPAn..51..145H}
Y.~{Han} and A.~{Yilmaz}, ``{Dynamic Routing for Navigation in Changing Unknown
  Maps Using Deep Reinforcement Learning},'' \emph{ISPRS Annals of
  Photogrammetry, Remote Sensing and Spatial Information Sciences}, vol.~51,
  pp. 145--150, Jun. 2021.

\end{thebibliography}

\end{document}